\documentclass[lettersize,journal]{IEEEtran}
\usepackage[english]{babel}
\usepackage{amssymb}
\usepackage{float}
\usepackage{booktabs}
\usepackage{algorithmic}
\usepackage{algorithm}
\usepackage{multirow} 
\usepackage{balance}
\usepackage{booktabs}
\usepackage[numbers,sort&compress]{natbib}

\usepackage[caption=false,font=normalsize,labelfont=sf,textfont=sf]{subfig}
\usepackage{amsmath}
\usepackage{graphicx}
\usepackage[colorlinks=true, allcolors=blue]{hyperref}

\title{Effective backdoor attack on graph neural networks in link prediction tasks}

\author{Haoyu Sun, Jiazhu Dai, Pin Wu
\thanks{Manuscript created June 2024, This paper was supported by Natural Science Foundation of Shanghai Municipality under Grant NO.22ZR1422600.(Corresponding author: Jiazhu Dai.)}
\thanks{Haoyu Sun and Jiazhu Dai are with the College of Computer Engineering and Science, Shanghai University, No.99, Shangda Road, Baoshan District, Shanghai, China. Email address: daijz@shu.edu.cn (Jiazhu Dai)
}}
\date{}
\begin{document}
\maketitle
\begin{abstract}
Graph Neural Networks (GNNs) are a class of deep learning models capable of processing graph-structured data, and they have demonstrated significant performance in a variety of real-world applications. Recent studies have found that GNNs models are vulnerable to a new type of threat called backdoor attacks, where the adversary can inject hidden backdoor into the GNNs models so that the attacked model performs well on benign samples, whereas it performs pre-specified malicious behavior such as misclassification to adversary-specified target category if the hidden backdoor is activated by samples with the attacker-defined pattern (such as subgraphs) called backdoor trigger. Backdoor attacks usually occur when the training process of GNNs models is not full controlled by users, such as training on third-party datasets or adopting third-party GNNs models, which poses a new and serious threat to GNNs models.

Currently, research on  backdoor attacks against GNNs models mainly focus on tasks such as graph classification and node classification, and backdoor attacks against link prediction tasks are rarely studied. In this paper, we propose a backdoor attack against the link prediction tasks based on GNNs and reveal that GNNs are vulnerable to such a threat. The proposed backdoor attack uses a single node as the backdoor trigger and injects backdoor into GNNs models through poisoning selected node pairs in the training graph. In the inference stage, the backdoor will be activated by simply linking the trigger node to the two end nodes of unlinked target node pairs in the input samples and the attacked GNNs models incorrectly predict link relationships between the unlinked target node pairs as linked. We demonstrate that this attack is effective and efficient by performing experimental evaluations on four popular models and four public datasets and comparing them with baseline, and the experimental results show that in the black-box scenario, this attack can achieve high attack success rates (more than 88 \%) with small model accuracy losses (less than 1\%) and small poisoning rates (about 1\%). 
\end{abstract}


\section{Introduction}

\IEEEPARstart{G}{raph-structured} data is ubiquitous in the real world, covering various application domains such as social networks \cite{borgatti2009network},urban transportation networks \cite{xie2019sequential}, knowledge graphs \cite{hogan2021knowledge}, and biological networks \cite{hirst1992prediction}. In order to efficiently process graph-structured data, various powerful graph neural networks(GNNs) have been proposed. GNNs adopt the message passing mechanism to update the representations of nodes by aggregating information from their neighbors. GNNs can be used for various graph related tasks such as graph classification \cite{xu2018powerful}, node classification \cite{kipf2016semi}, link prediction \cite{kipf2016variational,pan2018adversarially} etc. Link prediction aims to predict the existence of links between pairs of nodes in a graph, which is an important task of GNNs and has a wide range of applications such as friends recommendation in social networks \cite{daud2020applications}, prediction of missing triples in knowledge graphs \cite{nickel2015review}, and prediction of protein functions as well as interactions in biological networks \cite{qi2006evaluation}. 

GNNs have powerful performance in processing graph-structured data. However, recent studies \cite{xi2021graph,yang2022transferable,zheng2023motif,dai2023semantic,xu2021explainability,chen2022neighboring,chen2023feature,dai2023unnoticeable,zheng2023link,chen2023dyn,li2022backdoor} have shown that GNNs are vulnerable to a new security threat called backdoor attacks. In backdoor attacks, the attacker injects specific patterns (called backdoor triggers) in the training data (the process is called poisoning) and then embeds the backdoor in the GNNs models(called backdoored GNNs models) through training. In the inference phase, when the input samples contain the backdoor triggers, the backdoor in the models will be activated and perform the action specified by the attacker in advance. For example, the GNN models embedded with the backdoor misclassify input samples with triggers as attacker-specified target class label, whereas they perform normally when input samples that do not contain backdoor triggers. Backdoor attacks often occur when the training process of GNN models are uncontrolled, e.g. backdoor attacks can occur by using models trained by third parties or by using third-party data to train the models \cite{li2022backdoor}. Backdoor attacks are highly stealthy and they have posed serious security threat to GNNs.

Backdoor attacks have been extensively studied in computer vision and natural language processing \cite{gu2017badnets,chen2017targeted,dai2019backdoor,qi2021hidden}, while the threat of backdoor attacks on GNNs has not yet been fully explored. Current studies on backdoor attacks on GNNs have focused on the graph classification \cite{xi2021graph,yang2022transferable,zheng2023motif,dai2023semantic,xu2021explainability} and the node classification \cite{xu2021explainability,chen2022neighboring,chen2023feature,dai2023unnoticeable}, while there is few research on backdoor attacks against link prediction tasks.

To the best of our knowledge, the only existed backdoor attack studies on link prediction are LB \cite{zheng2023link} and DLB \cite{chen2023dyn}, where DLB is a backdoor attack method for link prediction on dynamic graphs, and LB is a backdoor attack method for link prediction on static graphs, which studies a similar problem to that of our work. Specifically, LB optimizes the structure and features of a random initial subgraph by using the gradient information of the target model, then obtains a dense subgraph as the trigger for the backdoor attack, and finally poisons the graph dataset with this subgraph, so as to embed the backdoor into the GNN models through training. Although LB achieves good performance, this method requires a large attack budget because LB uses the gradient information of the target model to optimize the subgraph, and the dense subgraph makes the trigger easy to be detected.

\begin{figure}[ht]
\centering
\includegraphics[width=1\linewidth]{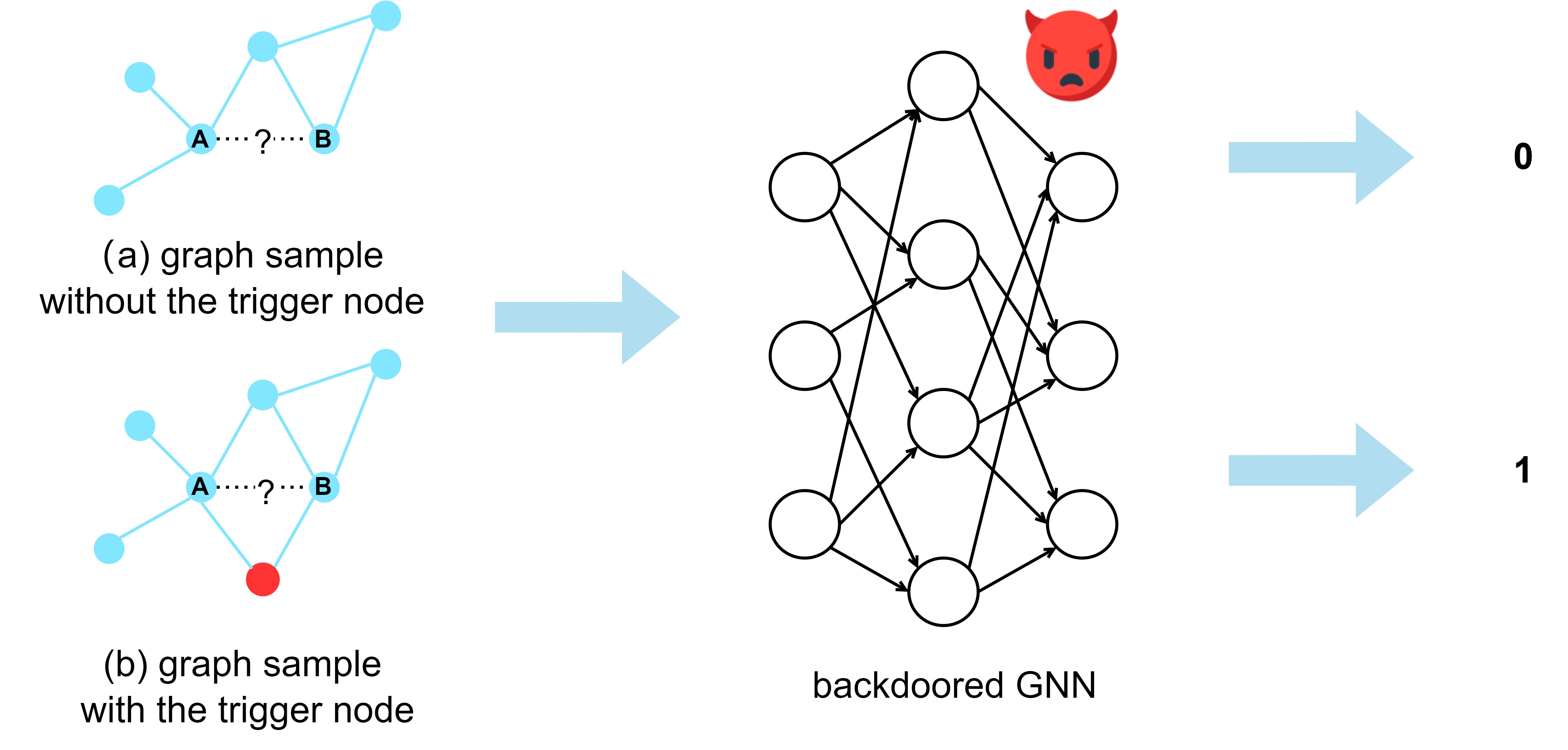}
\caption{\label{fig:introduction}Illustration of our backdoor attack against link prediction. The trigger is the red node and the node A and the node B is an unlinked target node pair. Part (a) is the input graph without the trigger. The backdoored model performs well for the input graph without the trigger and it predicts correctly that there is no link between the node A and the node B, which is represented by 0. The input graph in Part(b) is similar to that in Part(a) except that the former contains the trigger connecting to the node A and the node B respectively. The backdoor in the backdoored model is activated due to the presence of the trigger and it predicts incorrectly that there is a link between the node A and the node B for the input graph with the trigger, which is represented by 1.}
\end{figure}

In this paper, we investigate whether there are less costly and more stealthy backdoor attacks against link prediction task of GNNs. Specifically, we propose such a backdoor attack with a single node as the trigger, where the backdoored GNNs models incorrectly predict link relationships between two unlinked nodes as linked when the trigger occurs. Initially, a new node is created as the trigger, whose features are generated by counting the frequency of occurrence of each feature of all nodes and selecting the subset of features with fewer occurrences; Then some node pairs in the training graph are selected to inject the trigger for poisoning the training graph, and the backdoor is embedded in the model through the training process.  In the inference phase, the backdoor embedded in the model can be activated by simply connecting the trigger node to the two end nodes of the target unlinked node pairs in the test data, causing the backdoored model to incorrectly predict the unlinked two nodes as having a link relationship. The illustration of our backdoor attack is shown in Figure \ref{fig:introduction}. The left side of the figure shows the input graph and the red node is the backdoor trigger. Part (a) of Figure \ref{fig:introduction} is the input graph without the trigger, and node A and node B is an unlinked target node pair; The input graph in Part(b) is similar to that in Part(a) except that the former contains the trigger connecting to the node A and the node B respectively. When predicting the link relationship between nodes A and B in the input graph without trigger with the backdoored model, it works well and correctly predicts that there is no link between node A and node B, which is represented by 0, while for the input graph with the trigger as shown in Part (b) of Figure \ref{fig:introduction}, the backdoor in the backdoored model is activated due to the presence of the trigger and it predicts incorrectly that there is a link between the node A and the node B, which is represented by 1. We evaluate the performance of the proposed attack on four popular models and four benchmark datasets and compare the results with that of baseline to show that the attack is eﬀective and efficient, and the experimental results indicate that the attack success rate of the attack can reach more than 89 \% in the black-box scenario with the poisoning rates of about 1\%, and that the classification accuracy of the backdoored model on clean samples is close to that of clean model with the drop less than 1\%.

Our contributions in this paper are summarized as follows:

\begin{enumerate}
\item  We propose a backdoor attack against link prediction with GNNs, revealing that GNNs are vulnerable to this security risk. The backdoor attack has following three features: the attack achieves better stealthiness by using only one single node as the trigger, and it requires lower poisoning rate in training dataset ; The attack can be implemented in black box scenarios, where the attackers do not need to know the parameters of target GNN models or create surrogate models, so it is easier and more practical for attacker to launch the attack; The attack has lower computing cost without involving any computation of gradients.
\item We propose a fast and eﬀective trigger generation method, the core idea of which is to create the
features of the trigger based on the statistical results of each element in the feature vector of each node in the graph, so as to increase the discrepancy between the trigger node and normal nodes to improve the eﬀectiveness of the attack.
\item We propose a simple and efficient method for contaminating training data, which can achieve
better attack performance with lower poisoning rates by connecting the trigger node to the selected nodes pairs in the graph.
\item We evaluate the performance of our attack with four popular models and four benchmark datasets and compare the results with baseline. The experimental results show that our attack can achieve more than  89 \% attack success rates with poisoning rates of about 1 \% in the black-box scenario, while the classification accuracy of  the backdoored model on clean samples drops less than 1 \% in comparison with that of the clean models.
\end{enumerate}

The rest of the paper is organized as follows. Firstly, we introduce the background knowledge of GNNs, link prediction and backdoor attacks in Section \ref{section:2}; Then, we introduce the adversarial attacks against GNNs and further introduce present the state of art of backdoor attacks against GNNs in Section \ref{section:3}; Next, we describe in detail our proposed backdoor attack in Section \ref{section:4} and evaluate its performance in Section \ref{section:5};Finally, we conclude the paper in Section \ref{section:6}.

\section{Background}
\label{section:2}

\subsection{GNNs}

\noindent Given an undirected and unweighted attribute graph $G=(V,E,X)$, where $V=\{v_1,v_2,\cdots,v_N\}$ is the set of nodes, $E$ is the set of edges, $X=R^{N \times d}$ represents the nodes feature matrix, and the adjacency matrix $A \in R^{N\times N}$. For two nodes $v_i,v_j \in V$, if $(v_i,v_j)\in E$, it means that there exists an edge between $v_i$ with $v_j$ and $A_{ij}=1$, otherwise, $A_{ij}=0$.$ N=|V|$ is the total number of nodes, and $d$ is the number of node feature dimensions.

GNNs learn effective node representations $H^k\in R^{N\times F},k\in \{1,2,\cdots,K\}$ by combining graph structural information and nodes' features through the message propagation mechanism, where $H^k$ indicates the representation of the k-th layer and $H^0=X$. Formally, we can define the framework of GNNs as follows:
\begin{equation}
H^k=AGGREGATE^k \{A,H^{k-1},{\theta}^k\}
\end{equation}
where AGGREGATE function aggregates the neighboring features of each node in the graph to get new nodes’ features, $\theta$ is a learnable parameter. After obtaining the features of the nodes, they can be used for downstream tasks such as graph classification, node classification, link prediction, etc.

\subsection{Link prediction}

\noindent Link prediction is the prediction of potential links or missing links of node pairs in a network. Formally, given a graph $G=(V,E,X)$, all node pairs are represented by the set U in the graph $G$. $E_u=U-E$ denotes the links which are not present currently and there are potential links or missing links in $E_u$ in the graph. The purpose of link prediction is to predict these potential and missing links of node pairs. For clarity, in this paper, the node pair to be predicted for the existence or non-existence of link are all referred to as target node pair.

\subsection{Link prediction methods based on GNNs}

\noindent Link prediction methods based on GNNs have recently shown superior performance to traditional methods, which utilize nodes' features and structural information in the graph to learn the embedding of the nodes. The advantage of GNNs is that they can adaptively capture complex patterns and features in the graph without relying on artificially designed heuristic rules or assumptions. In the case of Graph Auto-Encoder (GAE) \cite{kipf2016variational}, GAE computes the node representation $z_i$ for each node by using two layers of graph convolutional networks (GCNs) \cite{kipf2016semi}
\begin{equation}
\begin{split}
Z=\tilde{D}^{-1/2} \tilde{A}\tilde{D}^{-1/2}\sigma&(\tilde{D}^{-1/2}\tilde{A}\tilde{D}^{-1/2}XW^0 ) W^1 \\
z_i &= Z_{i,:}
\end{split}
\end{equation}

where Z is the node embedding matrix output by GAE, and the i-th row of Z is the embedding of node $z_i$. $\tilde{A}=A+I$ is the adjacency matrix of the self-connection of the given undirected graph $G$, $\tilde{D}=\sum_{j}A_{ij}$ is the degree matrix, $W^0$,$W^1$represent the weight matrices of the first and second layers of the GCNs respectively. $\sigma (\cdot)$is the RELU activation function. Then an inner product decoder is used to reconstruct the edges in the graph, that is to predict whether a link exists between two nodes. Specifically, for any pair of nodes $i$ and $j$, the probability that the link exists between them is:
\begin{equation}
p_{\theta} (A_{ij}=1|X)=s(z_i^T z_j )
\end{equation}
where $s(\cdot)$ is the sigmoid function. The objective function of the GAE is to maximize the log-likelihood of all edges (positive samples) and randomly sampling an equal number of non-edges (negative samples) as positive samples:
\begin{equation}
\begin{split}
L(\theta)= &\sum_{(i,j)\in E^+}{logp_{\theta} (A_{ij}=1|X)} \\
&+ \sum_{(i,j)\in E^-}{log(1-p_{\theta}(A_{ij}=1|X))}
\end{split}
\end{equation}
where $E^+$ is the set of edges and $E^-$ is the set of non-edges. GAE can be trained by stochastic gradient descent optimization algorithm.

\subsection{Backdoor attacks}

\noindent Backdoor attacks are a type of poisoning attack against deep learning models, which aims to implant some hidden malicious functions, i.e., backdoors, in the target models by using poisoned samples with triggers during the models training process. In the inference phase, for input samples that contain trigger, the backdoor will be activated, causing the target models to produce incorrect prediction results, e.g., the input samples are incorrectly misclassified to the target class label specified by the attacker, whereas the backdoored model works normally for benign samples that do not contain trigger. Backdoor attacks were first proposed in the field of images \cite{gu2017badnets,chen2017targeted}. Gu et al. \cite{gu2017badnets} proposed the backdoor attack for the first time in the image domain, where the backdoored model is trained by poisoning some training samples using special markers as triggers, and in the inference phase, the backdoored model classifies the images with the triggers as the target label, and classifies the clean images normally. Subsequently, backdoor attacks in the textual domain have also been extensively studied \cite{dai2019backdoor,qi2021hidden}, e.g., Qi.et.al \cite{qi2021hidden} used syntactic structures as the triggers for textual backdoor attacks.

Backdoor attacks are highly stealthy and harmful because, in the absence of triggers, the performance and behavior of backdoored models are not significantly different from normal models and are not easily detected and defended against. Backdoor attacks can occur in many scenarios, such as when using datasets, platforms, or models provided by third parties for training or deployment. Backdoor attacks pose a serious threat to the security and trustworthiness of deep learning systems, and require sufficient attention and concern.

\section{Related works}
\label{section:3}

\noindent In this section, we briefly introduce the adversarial and backdoor attacks on GNNs.

\subsection{Adversarial attacks}

\noindent Although GNNs have achieved excellent performance in a variety of graph learning tasks and have been used in many applications, recent studies have shown that GNNs are equally susceptible to adversarial attacks \cite{dai2018adversarial,chen2018fast,zugner2018adversarial,wu2019adversarial,xu2019topology,zugner2020adversarial,dai2022targeted,tao2021single,wang2020evasion,wang2020scalable}. Depending on the stage of data perturbation that occurs during the attack, adversarial attacks can be divided into evasion attacks and poisoning attacks. In evasion attacks, we assume that the target model is a given already trained model that cannot be modified by the attacker, and the attacker makes the model predictions wrong by adding perturbations to the test samples. Dai et al \cite{dai2018adversarial} proposed reinforcement learning, genetic algorithm, and gradient based methods to optimize the perturbation of graph structure to implement adversarial attacks respectively. Chen et al \cite{chen2018fast} implemented the attack by modifying the links to change the embedding of the nodes using a fast gradient approach. These studies \cite{dai2018adversarial,chen2018fast,zugner2018adversarial,wu2019adversarial,xu2019topology} are practiced adversarial attacks by modifying the features of existing edges or nodes, which requires the attacker to have a high level of privilege, which is not practical in most of the cases in reality. Instead, some studies have explored more realistic approaches that do not modify the existing graph structure and node characteristics, but instead achieve this by injecting nodes and edges into the existing graph, which is more practical compared to the former. Wang et al \cite{wang2020scalable} reduced the classification accuracy of GCN by injecting malicious nodes into the graph. In poisoning attacks, the attacker perturbs the training set before the models is trained, causing a severe degradation in the performance of the trained models. For example, NETTACK \cite{zugner2018adversarial} initiates poisoning attacks by generating adversarial perturbations against node features and graph structures through the surrogate model; Metattack \cite{zugner2020adversarial} initiates poisoning attacks by modifying the structure of the graph through meta-learning.

\subsection{Backdoor attacks}

\noindent In the graph data field, backdoor attacks have been less studied, and current works focused on backdoor attacks against graph classification \cite{xi2021graph,yang2022transferable,zheng2023motif,dai2023semantic,xu2021explainability} and node classification \cite{xu2021explainability,chen2022neighboring,chen2023feature,dai2023unnoticeable}.  Xi et al. \cite{xi2021graph} proposed for the first time a backdoor attack method for GNNs that uses subgraphs as the triggers that can be dynamically customized for different graphs with different triggers to poison the data. Yang et al. \cite{yang2022transferable} revealed a transferable graph backdoor attack without a fixed trigger pattern, which implements black-box attacks on GNN by the attacking surrogate models. Zheng et al \cite{zheng2023motif} rethink the triggers from the perspective of motifs (motifs are frequent and statistically significant subgraphs  containing the rich structural information of graphs
), proposed motif-based backdoor attacks, and gave some in-depth explanations of backdoor attacks. Xu et al.\cite{xu2021explainability} select the optimal trigger position through interpretable methods to accomplish backdoor attacks against graph classification and node classification. 

The above studies mainly focus on graph classification and node classification, and the only current work on backdoor attacks under the context of link prediction are LB \cite{zheng2023link} and DLB \cite{chen2023dyn}. DLB investigated the link prediction backdoor attack on dynamic graphs, it generates different initial triggers by Generative Adversarial Network (GAN) \cite{goodfellow2020generative}, and then based on the gradient information of the attack discriminator in the GAN, selects some of the links of the initial triggers to form a trigger set to accomplish the backdoor attack.  LB is similar to our work, but LB uses subgraphs as triggers and generates triggers by using the gradient information of the target models. Compared to LB, Our proposed backdoor attack has following three different features: the attack achieves better stealthiness by using only one single node as the trigger, and it requires lower poisoning rate in training dataset; The attack can be implemented in black box scenarios, where the attackers do not need to know the parameters of target GNN models or create surrogate models, so it is easier and more practical for attacker to launch the attack; The attack has lower computing cost without involving any computation of gradients.

\section{Method}
\label{section:4}

\noindent In this section, we explain in detail how our attack is implemented. Table \ref{tab:1} summarizes the notions used in the following sections and their explanations.

\begin{table*}[!t]
\centering
\caption{\label{tab:1}The explanations of the notions}
\begin{tabular}{cc}
\toprule
Notations & Explanations \\
\midrule
$G=(V,E,X)$&Graph $G$ with edge set $E$, node set $V$ and feature matrix $X$ \\
$\lvert V \rvert$&The number of nodes \\
$f_c$&The clean GNN model for link prediction\\
$n_t$&The trigger node\\
$x_t$&The features vector of the trigger node\\
$f_b$&The backdoored GNN model for link prediction\\
$T$ &The linked state\\
$E_{ij}$&The target unlinked node pair without the trigger\\
$\widetilde{E_{ij}}$&The  target  unlinked node pair  with the trigger\\
$\alpha_j$&The frequency of occurrence of ”1” of the j-th feature\\
$M_k$&The top k indexes of dimension where the nodes with the least frequent occurrence\\
$k$&The number of modified trigger features\\
$p$&The poisoning rate\\
$Score$&The poisoned node pair scoring function\\
\bottomrule
\end{tabular}
\end{table*}

\subsection{Attack overview}

\noindent The backdoor attack proposed in this paper aims to the link prediction task under transductive learning, the goal of which is to incorrectly predict a link relationship between two unlinked target nodes. Formally, given a graph $G$ and a trigger $n_t$, the attacker generates the poisoned graph $\tilde{G}$ by embedding the trigger $n_t$ into the graph $G$. The backdoored model $f_b$ is obtained by training on $\tilde{G}$ and the clean model can be denoted as $f_c$. The backdoor attack target can be denoted as:
\begin{equation}\label{eq1}
\left\{
\begin{aligned}
f_b(\widetilde{E_{ij}})&=T\\
f_b(E_{ij})&=f_c(E_{ij})\\
\end{aligned}
\right.
\end{equation}
where $E_{ij}$ is the the target unlinked node pair without the trigger in graph G and $\widetilde{E_{ij}}$  denotes the target unlinked node pair injected with the trigger. The first formula of equation \ref{eq1} represents the effectiveness of the backdoor attack, which means that the backdoored GNNs model $f_b$ incorrectly predicts link state of the target unlink node pair 
 as the linked state $T$ due to the appearance of the trigger. The second formula of equation \ref{eq1} represents the evasiveness of the backdoor attack,  which means that the backdoored GNN model works well for  the target unlinked node pair $E_{ij}$ without the trigger and its predicting accuracy is as close as possible to that of the corresponding clean model.

\begin{figure*}[!t]
\centering
\includegraphics[width=1\linewidth]{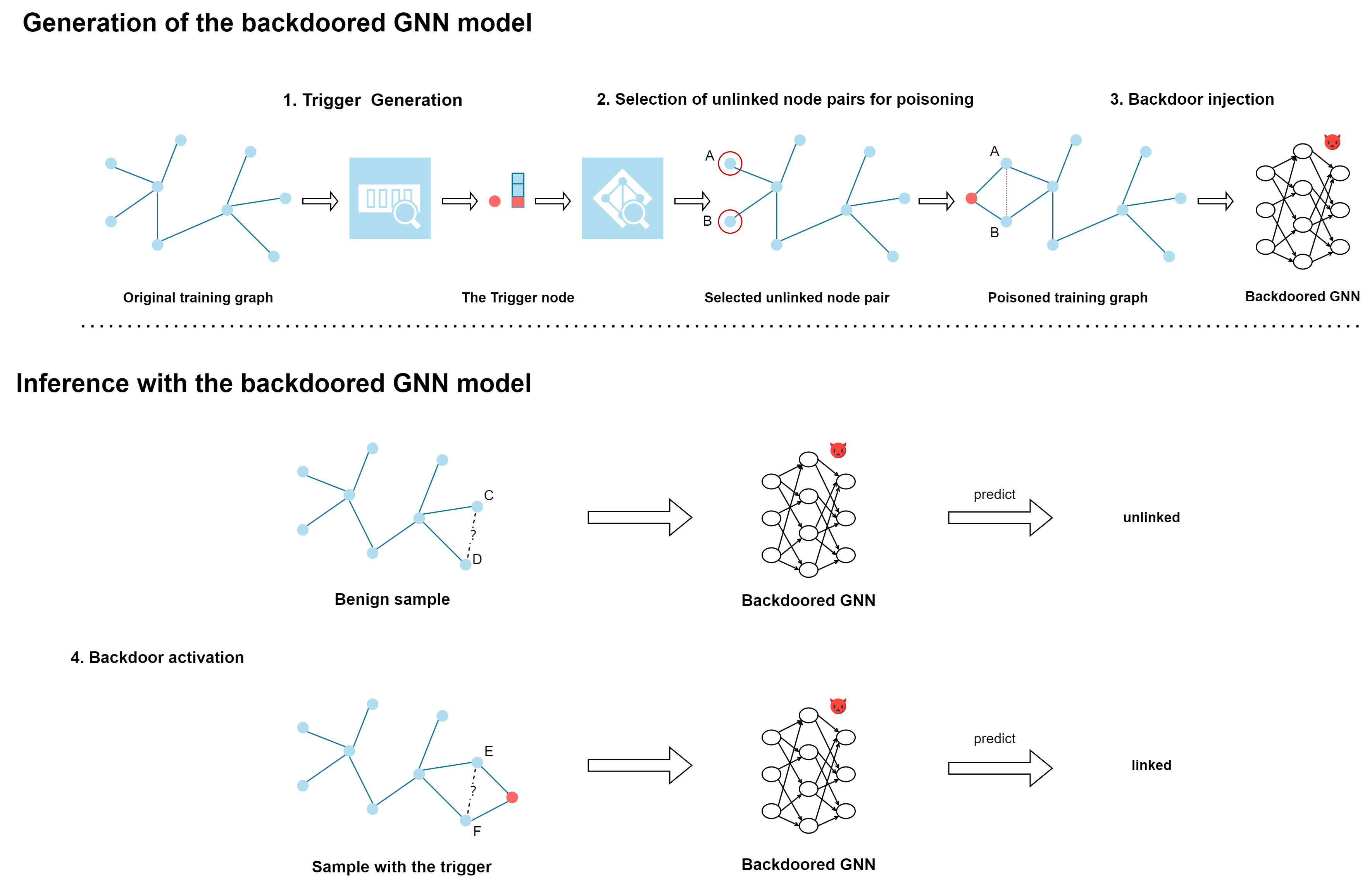}
\caption{\label{fig:attack}The proposed backdoor attack framework. The backdoor attack consists of four steps:1. The attacker generates the trigger node, as shown in red dot in the figure, the features of which are got by selecting the top k features with the least occurrence frequency, based on statistical information of the features of all nodes in the original training dataset. 2. The attacker selects unlinked node pairs for poisoning from the training graph through a score function, as shown in the figure, node A and node B are one of such selected unlinked node pairs. 3. The attacker generates the poisoning training dataset by injecting the trigger into the above selected node pairs (i.e. connect the trigger node to two end nodes of the node pairs, for example, connect the red trigger node to node A and node B in the figure) and change the unlinked state of the selected node pairs to linked state. Then the poisoning training dataset is used to train the GNN models to embed the backdoor in the models, which is called backdoored GNNs models. 4. In the inference phase of the backdoored GNN models, if the unlinked node pairs are connected to the trigger node , such as the node E and the node F in the sample with the trigger in the figure, they will activate the backdoor in the models and will be predicted incorrectly to be linked, while for the unlinked node pairs without the trigger node, such as the node C and node D in the benign sample in the figure, they will be predicted correctly to be unlinked.}
\end{figure*}

Figure \ref{fig:attack} depicts the process of this backdoor attack. It includes the following steps:

\begin{enumerate}
\item \textbf{Trigger generation}: the features of the trigger node are generated by selecting the top k features with the least occurrence frequency, based on statistical information  of  the features of all nodes in the training data, as detailed in Section \ref{section:4.2} below;
\item \textbf{Selection of node pairs for poisoning}: a scoring function is proposed to select suitable node pairs in the training graph, which will be injected with the trigger node in the next step to generated poisoning training dataset. as detailed in Section \ref{section:4.3} below;
\item \textbf{Backdoor injection}: The poisoned training dataset is generated by injecting the trigger node  into the selected node pairs from step 2 above and changing the link state of the selected unlinked node pairs to be linked. The poisoned training dataset is then used to train the GNN models to inject the backdoor, as detailed in Section \ref{section:4.4} below;
\item \textbf{Backdoor activation}: in the inference phase, when the trigger node is connected to the end nodes of target unlinked node pair respectively, the backdoor in the backdoored model will be activated and it will incorrectly predict that there is a link between these two unlinked nodes. And when the trigger is not present, the backdoored model work well.
\end{enumerate}

The backdoor attack proposed in this paper has the following assumptions:

\begin{enumerate}
\item We focus on link prediction under transduction learning. The goal of this backdoor attack is to incorrectly predict unlinked node pairs as linked.
\item The attacker can access to the training data, including the topology and features of the nodes, and the attacker can add additional edges and nodes to the graph to generate poisoning data.
\item The feature matrix of the graph is binary.
\end{enumerate}

\subsection{Trigger Generation}
\label{section:4.2}

\noindent In this paper, we use a single node as the trigger and generate its feature vector by the following two steps:

1.Identify top k indexes of dimensions where element “1” occurs least frequently in the binary feature vectors of all nodes in the training dataset.

Zheng et al. found in their research on motif-Backdoor \cite{zheng2023motif} that using non-existent or infrequent substructures as the triggers is more effective in implementing backdoor attacks for GNNs. Inspired by this method, we use a single node as the trigger and distinguish the trigger node from other normal nodes in the training dataset in terms of the feature vector, so as to form a strong association between the trigger node and the linked state in the training process.

Specifically, in an undirected unweighted feature graph $G=(V,E,X)$, where $V=\{v_1,v_2,\cdots,v_N\}$ is the set of nodes, $E$ is the set of edges, $X=R^{N\times d}, j\in \{1,2,\cdots,d\}$ represents features matrix of the nodes, the rows of which are d-dimensional feature vectors of the corresponding nodes.
in order to determine the indexes of dimension where element “1” occurs least frequently in the binary feature vectors of all nodes in the training dataset. we compute the $L_1$ norm of each column vector of the above matrix $X$ as follows:

\begin{equation}\label{eq2}
\alpha_j=\Vert X_{:,j} \Vert_1,j \in \{1,2,\cdots,d\}
\end{equation}
where $X_{:,j}$ represents the d-dimensional column vectors in the node feature matrix. Since we only consider the case where the features of the nodes in the graph are binary, the above formula calculates the frequency of occurrence of "1" in each dimension of the feature vectors of all nodes, which is denoted as $\{ \alpha_1, \alpha_2,\cdots,\alpha_d\}$. We fututer select the smallest $k$ values among them, denoted as $\{\alpha_{m_1}, \alpha_{m_2},\cdots,\alpha_{m_k}\}$, where subscripts set $M_k=\{m_1,m_2,\cdots,m_k \}\subseteq\{1,2,3\dots,d\}$ stands for the top $k$ indexes of dimension where element “1” occurs least frequently in the binary feature vectors of all nodes in the training dataset. $k$ is called trigger size which is the number of modified elements in the feature vector of the trigger node ,as described in the following step.

2. Modify the value of the corresponding dimension in the feature vector of the trigger node to "1" based on the above dimension indexes $M_k$.

At first, all the elements in the feature vector of the trigger node are initially set to "0", and then according to the above $k$ dimension indexes, the value of the corresponding dimension in the feature vector of the trigger node are modified to "1" . We denote the trigger node by $n_t$ and its i-th feature value by $x_t^i$, the above operation can be expressed as:
\begin{equation}\label{eq3}
x_t^i=\left\{
\begin{aligned}
1, \qquad i\in M_k\\
0, \qquad i\notin M_k\\
\end{aligned}
\right.
\end{equation}

After the above two steps, we have generated the trigger node, which will be used to poison the training dataset in the following step.

\subsection{Selection of node pairs for poisoning}
\label{section:4.3}

\noindent After generating the trigger node features, the next step is to select the suitable node pairs to inject the trigger node to generate effective poisoning training data. 

Considering that the poisoned training dataset is generated by connecting the trigger node with the end nodes of the selected unlinked node pairs in the original training graph. The links between the trigger node and the end nodes of these unlinked node pairs may affect learning the features of the trigger node by the GNN model. Due to the difficulty of GNNs in extracting hidden features from sparse node features (sparse meaning that the number of elements with value 1 in the node’s feature vector is far less than that of elements with value 0)\cite{ding2023hyperformer,ding2019feature}, these node pairs with sparse node features will have less impact on learning the features of the trigger node by the GNN model . Therefore, we select node pairs with sparse features to inject the trigger node so as to enhance the influence of the trigger node in the backdoor attack and form a strong association between the trigger node and linked state in the training process. The scoring function to select unlinked node pairs with sparse features to inject the trigger node is defined as follows:

\begin{equation}\label{eq4}
Score(v_i,v_j) = \Vert X_{v_i} \Vert_1 + \Vert X_{v_j} \Vert_1
\end{equation}
where, $(v_i,v_j )$ is an unlinked node pair in the training data, $\Vert X_{v_i} \Vert_1$ and $\Vert X_{v_j} \Vert_1$ represent the number of non-zero elements of the feature vectors of nodes $v_i$ and $v_j$, respectively. As can be seen, the sparser the features of the node pairs, the smaller the scores.

For all the unlinked node pairs in the training data, we compute their value with the above score function respectively, and select $q$ unlinked node pairs with the smallest score values as the target node pairs to inject the trigger, where, $q = p\cdot |V|$, $p$ is the poisoning rate, which represents the ratio of the number of poisoned node pairs to the total number of nodes in the training graph.

\subsection{Backdoor injection}
\label{section:4.4}
\noindent After generating the trigger node and selecting the node pairs for poisoning, the next step is to generate poisoning training dataset $\widetilde{G}$ by injecting the trigger node into the selected node pairs(i.e. connect the trigger node with two end nodes of the node pairs) and changing the unlinked state of the node pairs to be linked. Then the poisoning training dataset is used to train the GNN models to embed the backdoor in the models. The overall algorithm is shown in algorithm \ref{alg:1}.

\subsection{Backdoor Activation}

\noindent In the inference phase, if the trigger node is connected to two end nodes of unlinked target node pair respectively, the backdoor in the backdoored models will be activated and they will incorrectly predict that there is a link in the unlinked node pair, while for unlinked node pairs without the trigger node, the backdoored models work normally.

\begin{algorithm}[htbp]
    \caption{Algorithm of our backdoor attack}
    \label{alg:1}
    \renewcommand{\algorithmicrequire}{\textbf{Input:}}
    \renewcommand{\algorithmicensure}{\textbf{Output:}}
    \begin{algorithmic}[1]
        \REQUIRE $G=(V,E,X)$,poisoning rate $p$  

        \FOR{$i=1$ to $N$}
            \STATE Calculate $\alpha_i$ of each feature dimension by equation \ref{eq2}
        \ENDFOR

        \STATE Get the dimension indexes $M_k$ by selecting the k dimensions with minimum $\alpha_i$

        \STATE Generate the features $x_t$ of the trigger node by equation \ref{eq3}

        \STATE Calculate the score of node pairs by equation \ref{eq4}
        
        \STATE select unlinked node pairs for poisoning according the above score

        \STATE Generate the poisoning graph by injecting the trigger node into the selected unlinked node pairs and changing the link state of the unlinked node pairs to be linked

        \STATE Train the model on poisoning graph $\widetilde{G}$ to obtain $f_b$

        \ENSURE backdoored model $f_b$  and trigger $n_t$    
    \end{algorithmic}
\end{algorithm}

\section{Experiment}
\label{section:5}

\noindent The following four experiments have been conducted to evaluate the effectiveness of the proposed backdoor attack. Firstly, we generated the backdoored GNN models with different poisoning rates and tested the attack success rate of these backdoored models on samples with trigger as well as their prediction accuracy on clean samples respectively to evaluate the effectiveness and stealthiness of the proposed backdoor attack. as described in Section \ref{section:5.2}. Secondly, we compared these results with the baseline to assess the performance of the proposed backdoor attack, as described in Section \ref{section:5.3}. Next we tested the impacts of different trigger sizes on the attack success rates of the backdoored models and their prediction accuracy on clean samples respectively, as described in Section \ref{section:5.4}. Finally we conducted the ablation studies to confirm the eﬀectiveness of the score function used to select node pairs for poisoning on the performance of the proposed backdoor attack, as described in Section \ref{section:5.5}.

\subsection{Experiment setting}
\subsubsection{Datasets}

We evaluate the proposed backdoor attack in four graph benchmark datasets: Cora\cite{mccallum2000automating}, CiteSeer\cite{giles1998citeseer}, CS\cite{mcauley2015image}, and Physcics\cite{mcauley2015image}. They are all undirected graphs composed of literature citation networks, where nodes represent papers, edges represent citation relations, node features are word vectors, and node labels are paper topics. Table \ref{tab:2} summarizes the basic statistics of these four graph databases.

\begin{table}[htbp]
\caption{\label{tab:2}The dataset statistics}
\centering
\begin{tabular}{ccccc}
\toprule
Datasets&Nodes&Edges&Classed&Features\\
\midrule
Cora&2708&5429&7&1433\\
CiteSeer&3327&4732&6&3703\\
CS&18333&327576&15&6805\\
Physcics&34493&247862&5&8415\\
\bottomrule
\end{tabular}
\end{table}

\subsubsection{Link prediction models}

In order to evaluate the effectiveness of the proposed backdoor attack, we use four popular GNN models for link prediction, that is, Graph Auto-Encoder (GAE)\cite{kipf2016variational}, Variational Graph Auto-Encoder (VGAE) \cite{kipf2016variational}, Adversarial Regularized Graph Auto-Encoder (ARGA) \cite{pan2018adversarially}, and Adversarial Regularized Variational Graph Auto-Encoder (ARVGA)\cite{pan2018adversarially}. We briefly describe these models as follows:

\textbf{GAE}\cite{kipf2016variational}: GAE is the auto-encoder based graph embedding method that uses GCNs as an encoder to map the nodes in the graph to a low dimensional vector space and then uses inner product as a decoder to reconstruct the adjacency matrix of the graph.

\textbf{VGAE}\cite{kipf2016variational}: VGAE is a graph embedding method based on variational auto-encoder, which differs from GAE in the encoder part. Instead of using GCNs directly to get the node vector representations, VGAE uses two GCNs to get the mean and variance of the node vector representations respectively, and then samples the node vector representations from a Gaussian distribution. This increases the diversity and robustness of the node vector representation.

\textbf{ARGA}\cite{pan2018adversarially}: ARGA is a graph embedding method based on adversarial regularization, which differs from GAE and VGAE in the regularization part.Instead of using KL scatter to constrain the node vector representations to obey a certain distribution, ARGA uses a discriminator to judge whether the node vector representations come from the real data distribution. This can make the node vector representation more consistent with the real data distribution, and at the same time avoid the loss of information caused by the KL scatter.

\textbf{ARVGA}\cite{pan2018adversarially}: ARVGA is also a graph embedding method based on adversarial regularization, which combines the techniques of variational auto-encoder and adversarial regularization. ARVGA employs two GCN networks to generate the mean and variance of the node vector representations, respectively, and uses a discriminator through adversarial training to improve the realism of the node representations.

\subsubsection{Parameter setting}

We construct GAE, VGAE, ARGA and ARVGA models for link prediction and adopt the same settings as those in the corresponding original papers. We construct two-layer GCNs with a 32-dim hidden layers and 16-dim hidden layers in all. For ARGA and ARVGA,  the discriminators are built with two hidden layers (16-neuron, 64-neuron respectively). In the experiment, we repeated the experiment five times and averaged the results.

\subsubsection{Metrics}
\label{section:5.1.4}

To evaluate the effectiveness and evasiveness of the backdoor attack proposed in this paper, we use the following metrics.

\begin{enumerate}
\item Attack success rate (ASR) refers to the ratio of the number of the attack node pairs (i.e. target unlinked node pairs with the trigger) predicted to be linked to the total number of the attack node pairs. It is as follows
\begin{equation}
ASR=\frac{N_{suc}}{N_{att}}
\end{equation}

where $N_{suc}$ is the number of the attack node pairs predicted to be linked, $N_{att}$ is the total number of attack node pairs.
\item Benign performance drop (BPD): To evaluate the evasiveness of our attack, we utilize the area under the curve, AUC\cite{huang2005using}, to represent the accuracy of the link prediction model. If, among $n$ independent comparisons, the number of times that the existing link gets a higher score than the nonexistent link is $n^{'}$, and the number of times they get the same score is $n''$, then the AUC is defined as
\begin{equation}
    AUC = \frac{n^{'}+0.5 n^{''}}{n}
\end{equation}

We then use the Benign Performance Difference (BPD), which measures the difference in prediction accuracy between the clean GNN model and the backdoored GNN model on clean graph. The BPD is defined as
\begin{equation}
    BPD=AUC_{c}-AUC_{b}
\end{equation}

where $AUC_c$ and $AUC_b$ represent the accuracy of clean model and backdoored model respectively. Lower BPD represents better performance, which indicates that backdoored model and clean model are closer to each other in terms of accuracy on clean samples, and the backdoor attack achieves more stealthy.
\item Poisoning rate (p): The poisoning rate $p$ represents the ratio of the number of target unlinked node pairs embedded with the trigger node to the total number of nodes in the graph. The lower the poisoning rate, the easier and more stealthier the backdoor attack will be.
\item Trigger size(k): The $k$ represents the number of elements in the feature vector of the trigger node that have been changed from an initial value of 0 to a value of 1 in the process of generating the trigger node. The trigger size is defined as follows
\begin{equation}
    k=\lambda d
\end{equation}

where $k$ is the trigger size, and $d$ is the total number of the trigger node feature dimensions, and $\lambda$ is the budget which represents the percentage of the modified  features among all features in the  trigger node.
\end{enumerate}

\subsubsection{Baseline}
In order to evaluate the proposed backdoor attack, we compare it with the only one state-of-the-art baseline LB\cite{zheng2023link} which is a backdoor attack method for link prediction tasks on static graphs, using subgraphs as triggers.

\subsection{Testing the impact of different poisoning rates on ASR and BPD}
\label{section:5.2}

\noindent We test the ASR and BPD of the backdoored models generated from the four GNN models GAE, VGAE, ARGA and ARVGA with different poisoning rates on four datasets to evaluate the impact of the poisoning rate on the effectiveness and evasiveness of the backdoor attack method proposed in this paper, where the poisoning rates in each dataset of the experiments are 0.2\%, 0.5\%, 1\%, 2\%, 5\%, respectively. The results of the experiments are shown in the table \ref{tab:3}.

\begin{table*}
\begin{center}
\caption{\label{tab:3}The impact of the poisoning rate on ASR and BPD.}
\begin{tabular}{@{}cccccccccc@{}}
\toprule
\multirow{2}{*}{Datasets} & \multirow{2}{*}{p(\%)} & \multicolumn{4}{c}{ASR(\%)}                                                                       & \multicolumn{4}{c}{BPD(\%)}                                                                       \\ \cmidrule(l){3-10} 
                         &                    & \multicolumn{1}{c}{GAE} & \multicolumn{1}{c}{VGAE} & \multicolumn{1}{c}{ARGA} & \multicolumn{1}{c}{ARVGA} & \multicolumn{1}{c}{GAE} & \multicolumn{1}{c}{VGAE} & \multicolumn{1}{c}{ARGA} & \multicolumn{1}{c}{ARVGA} \\ \cmidrule(r){1-10}
\multirow{5}{*}{Cora}    & 0.2         &95.39   &94.59    &91.55   &74.82   &-0.31  &0.32   &-0.64  &-0.01                     \\
                         & 0.5         &98.84   &98.43    &96.32   &79.76   &-0.64  &0.28   &-0.62  &-0.16                     \\
                         & 1           &99.00   &99.13    &97.53   &88.16   &-1.30  &0.17   &-0.71  &-0.27                    \\
                         & 2           &99.50   &99.33    &97.87   &91.79   &-1.24  &-0.18  &0.62   &-0.62                     \\
                         & 5           &99.76   &99.61    &98.76   &94.72   &-0.80  &0.10   &-0.62  &-0.27                     \\ \cmidrule(l){1-10}
\multirow{5}{*}{CiteSeer}& 0.2         &98.45   &98.47    &98.41   &95.85   &0.30   &0.65   &-0.43  &0.42                      \\
                         & 0.5         &99.42   &98.63    &96.44   &97.69   &-0.28  &-0.38  &-0.40  &0.19                     \\
                         & 1           &99.69   &99.85    &99.81   &98.88   &0.32   &0.02   &-0.40  &0.37                       \\
                         & 2           &99.77   &99.87    &97.97   &99.10   &0.48   &-0.34  &-0.48  &0.49                       \\
                         & 5           &99.90   &99.95    &99.95   &99.49   &-0.78  &-0.49  &-0.24  &0.54                       \\ \cmidrule(l){1-10}
\multirow{5}{*}{CS}      & 0.2         &94.12   &96.07    &94.59   &87.51   &0.50   &0.13   &-0.02  &0.25                       \\
                         & 0.5         &97.38   &98.03    &96.85   &91.00   &0.76   &0.08   &-0.02  &0.25                       \\
                         & 1           &98.36   &98.63    &97.72   &93.72   &0.59   &0.03   &-0.03  &0.30                       \\
                         & 2           &99.77   &98.87    &99.97   &99.10   &0.46   &-0.37  &-0.48  &0.49                        \\
                         & 5           &99.91   &99.95    &99.95   &99.49   &-0.73  &-0.50  &-0.24  &0.54                      \\  \cmidrule(l){1-10}
\multirow{5}{*}{Physics} & 0.2         &98.65   &99.19    &93.93   &93.19   &-0.44  &-0.04  &0.11   &-0.65                       \\
                         & 0.5         &99.81   &98.96    &95.86   &96.28   &-0.41  &-0.01  &0.11   &-0.56                       \\
                         & 1           &99.94   &99.73    &98.09   &97.10   &-0.44  &-0.05  &0.10   &-0.59                      \\
                         & 2           &99.95   &99.78    &99.02   &98.12   &-0.46  &0.01   &0.10   &-0.56                       \\
                         & 5           &99.97   &99.88    &99.31   &98.51   &-0.43  &0.06   &0.08   &0.47                       \\ \cmidrule(l){1-10}

\end{tabular}
\end{center}
\end{table*}

\begin{figure*}
\centering
\includegraphics[width=1\linewidth]{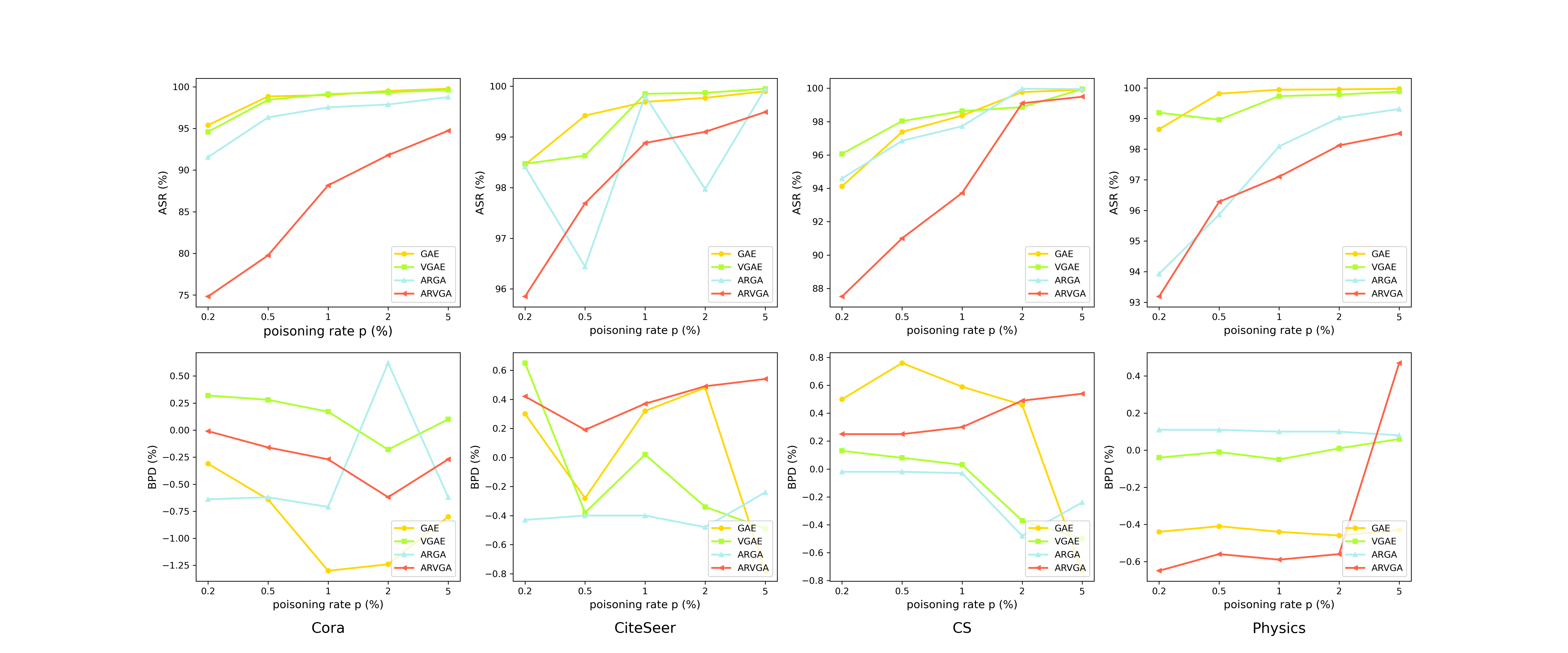}
\caption{ The impact of the poisoning rate on ASR and BPD.}
\label{fig:rate}
\end{figure*}

Figure \ref{fig:rate} is a visualization of the above experimental results, in which the vertical coordinates of the top four charts are the ASR, the vertical coordinates of the bottom four charts are the BPD, and the horizontal coordinates of all charts are the poisoning rates. From figure \ref{fig:rate}, we can observe that as p increases, the ASRs of the backdoored models increases over the four datasets. When the poisoning rate is increased to 2\%, our attack achieves excellent attack performance on all datasets. Taking the Cora dataset as an example, when $p=0.2\%$, the ASRs of the four backdoored models GAE, VGAE, ARGA and ARVGA are 95.39\%, 94.59\%, 91.55\%, 74.82\% respectively, and their ASRs are increased to 99.50\%, 99.33\%, 97.87\%, 91.79\% respectively when $p=2\%$. Thus, the proposed backdoor attack can achieves good attack performance with a small poisoning rate, which ensures that our attack is effective and stealthy.

From figure \ref{fig:rate}, we can also observe that the BPD of the four backdoored models in the four datasets fluctuates somewhat with $p$, but the fluctuation is overall less than 1\%. The results indicate that the backdoored models have close prediction accuracy to that of clean models on benign samples, which makes the backdoored models very stealthy.

\subsection{Comparison with baseline}
\label{section:5.3}
\noindent In order to evaluate the proposed backdoor attack,we compare it with the only one state-of-the-art baseline LB\cite{zheng2023link} which is a backdoor attack method for link prediction tasks on static graphs ,using subgraphs as triggers. As the baseline has conducted backdoor attack performance tests on models(GAE, VGAE, ARGA and ARVGA) across three datasets(Cora, CiteSeer, CS), and also performed a backdoor attack performance test on model GAE on the large scale dataset Physics, we have compared the performance of our proposed backdoor attack method with the baseline accordingly.

\noindent 1. Comparison of the performance of the proposed backdoor attack  with that of the baseline on four models(GAE, VGAE, ARGA and ARVGA)  across three datasets(Cora, CiteSeer, CS).

We test the performance of our backdoor attack on three datasets (Cora, CiteSeer, and CS) for four link prediction models (GAE, VGAE, ARGA and ARVGA) and compared it with the experimental results of baseline, where, the experimental results of baseline come from the original paper\cite{zheng2023link} and they were obtained with a poisoning rate of 10\% poisoning rate. Since our method requires only small poisoning rates to achieve attack success rate, we choose the experimental results with the 1\% poisoning rate to compare with the baseline, as shown in table \ref{tab:4}.

\begin{table}[htbp]
\centering
\caption{\label{tab:4} Comparison with the performance of the baseline on four models across three datasets.}
\begin{tabular}{@{}cccccc@{}}
\toprule
\multirow{2}{*}{Datasets} & \multirow{2}{*}{models} & \multicolumn{2}{c}{ASR(\%)}                                 & \multicolumn{2}{c}{BPD(\%)}                                                                       \\ \cmidrule(l){3-6} 
                         &                    & \multicolumn{1}{c}{baseline} & \multicolumn{1}{c}{ours}  & \multicolumn{1}{c}{baseline} & \multicolumn{1}{c}{ours}  \\ \cmidrule(r){1-6}
\multirow{4}{*}{Cora}    &GAE          &81.25   &99.00     &2.82  &-1.30                     \\
                         &VGAE         &83.22   &99.13     &4.65  &0.17                     \\
                         &ARGA         &92.56   &97.53     &1.08  &-0.71                     \\
                         &ARVGA        &85.25   &88.16     &2.49  &-0.27                     \\ \cmidrule(l){1-6}
\multirow{4}{*}{CiteSeer}&GAE          &82.25   &99.69     &3.13  &0.32                      \\
                         &VGAE         &89.59   &99.85     &8.51  &0.02                      \\
                         &ARGA         &98.78   &99.81     &4.97  &-0.40                       \\
                         &ARVGA        &81.55   &98.88     &1.60  &0.37                       \\ \cmidrule(l){1-6}
\multirow{4}{*}{CS}      &GAE          &50.61   &98.36     &1.13  &0.59                        \\
                         &VGAE         &67.26   &98.63     &0.95  &0.03                       \\
                         &ARGA         &63.52   &97.72     &0.32  &-0.03                       \\
                         &ARVGA        &68.67   &93.72     &2.86  &0.30                       \\  \cmidrule(l){1-6}
\end{tabular}
\end{table}

From the experimental results in table \ref{tab:4}, the performance of the proposed backdoor attack against four models is better than that of the baseline on all three datasets, with four backdoor models achieving ASRs of over 88\% and BPD fluctuation of less than 1.5\% on the three datasets. Taking the Cora dataset as an example, our method achieved 99.00\%, 99.13\%, 97.53\%, and 88.16\% ASRs respectively on the four backdoored models of GAE, VGAE, ARGA and ARVG, which are higher than the baseline's 81.25\%, 83.22\%, 92.56\%, and 85.25\%, respectively. All of our results on the Citeseer dataset achieve an ASR of more than 99\%. On the CS dataset, the ASRs of baseline drops dramatically, while our method still achieves an over 98\% ASRs at CS.

\noindent 2. Comparison of the performance of the proposed backdoor attack  with that of the baseline on GAE using large -scale dataset.

Considering that graph data in the real world are complex, the baseline evaluates its performance on the large-scale dataset physcics. It only tested the performance of the backdoored GAE model with the poisoning rate of 10\% on the dataset physics. So we choose from table \ref{tab:3} the results of backdoored GAE model with a poisoning rate of 1\%  on the dataset physics to compare with it. The comparison is shown in table \ref{tab:5}.

From table \ref{tab:5}, we can see that the performance of the proposed backdoor attack exceeds that of the baseline, with a 60\% increase in ASR and a 0.78\% reduction in BPD.

\begin{table}[htbp]
\centering
\caption{\label{tab:5}Comparison with the baseline on the large-scale dataset}
\begin{tabular}{ccccc}
\toprule
Method&ASR(\%)&BPD(\%)\\
\midrule
baseline&39.02&0.78\\
ours&99.94&-0.44\\
\bottomrule
\end{tabular}

\end{table}

\begin{table*}[ht]
\centering
\caption{\label{tab:6}The impact of trigger size on ASR and BPD.}
\begin{tabular}{@{}cccccccccc@{}}
\toprule
\multirow{2}{*}{Datasets} & \multirow{2}{*}{$\lambda$(\%)} & \multicolumn{4}{c}{ASR(\%)}                                                                       & \multicolumn{4}{c}{BPD(\%)}                                                                       \\ \cmidrule(l){3-10} 
                         &                    & \multicolumn{1}{c}{GAE} & \multicolumn{1}{c}{VGAE} & \multicolumn{1}{c}{ARGA} & \multicolumn{1}{c}{ARVGA} & \multicolumn{1}{c}{GAE} & \multicolumn{1}{c}{VGAE} & \multicolumn{1}{c}{ARGA} & \multicolumn{1}{c}{ARVGA} \\ \cmidrule(r){1-10}
\multirow{5}{*}{Cora}    & 0.2         &94.31    &96.14    &84.01  &66.48   & -1.27  &0.18  &-0.69  &-0.28                   \\
                         & 0.5         &98.48    &98.63   &93.52   &75.18   &-1.29   &0.21  &-0.70  &-0.28                    \\
                         & 1           &99.98   &99.13    &97.53   &88.15   &-1.28   &0.17  &-0.71  &-0.27                     \\
                         & 2           &99.44   &99.55    &98.83   &96.42   &-1.32   &0.20  &-0.72  &-0.22                     \\
                         & 5           &99.66   &99.57    &99.46   &99.22   &-1.38   &0.22  &-0.72  &-0.28                    \\ \cmidrule(l){1-10}
\multirow{5}{*}{CiteSeer}& 0.2         &99.01   &99.13    &97.37   &69.91   &0.76    &-0.20 &-0.38  &0.36                     \\
                         & 0.5         &99.72   &99.66    &99.49   &90.04   &0.71    &-0.08 &-0.35  &0.28                     \\
                         & 1           &99.69   &99.85    &99.81   &98.88   &0.32    &0.02  &-0.40  &0.37                       \\
                         & 2           &99.95  &99.93     &99.92   &99.63   &0.52    &0.11  &-0.39  &0.45                       \\
                         & 5           &99.87  &99.92     &99.92   &99.91   &0.68    &0.07  &-0.39  &0.39                       \\ \cmidrule(l){1-10}
\multirow{5}{*}{CS}      & 0.2         &89.96  &90.64     &84.63   &65.42   &0.60    &0.04  &0.02   &0.30                        \\
                         & 0.5         &96.91   &96.92    &94.92   &82.51   &0.62    &0.05  &-0.03  &0.30                       \\
                         & 1           &98.36   &98.63    &97.72   &93.72   &0.59    &0.03  &-0.03  &0.30                      \\
                         & 2           &99.94   &99.94    &99.92   &99.63   &0.40    &0.09  &-0.39  &0.45                        \\
                         & 5           &99.87   &99.92    &99.98   &99.63   &0.40    &0.09  &-0.39  &0.45                       \\  \cmidrule(l){1-10}
\multirow{5}{*}{Physics} & 0.2         &97.99   &97.33    &85.50   &68.61   &-0.41   &-0.04 &0.10   &-0.59                      \\
                         & 0.5         &99.52   &98.95    &96.40   &89.16   &-0.40   &-0.04 &0.10   &-0.59               \\
                         & 1           &99.94   &99.73    &98.09   &97.10   &-0.44	&-0.05 &0.10   &-0.59                \\
                         & 2           &99.94   &99.95    &98.76   &98.77   &-0.47	&-0.03 &0.11   &-0.59                \\
                         & 5           &99.98   &99.99    &99.85   &99.82   &-0.43	&-0.05 &0.11   &-0.59                \\ \cmidrule(l){1-10}

\end{tabular}
\end{table*}

\begin{figure*}[ht]
\centering
\includegraphics[width=2.0\columnwidth]{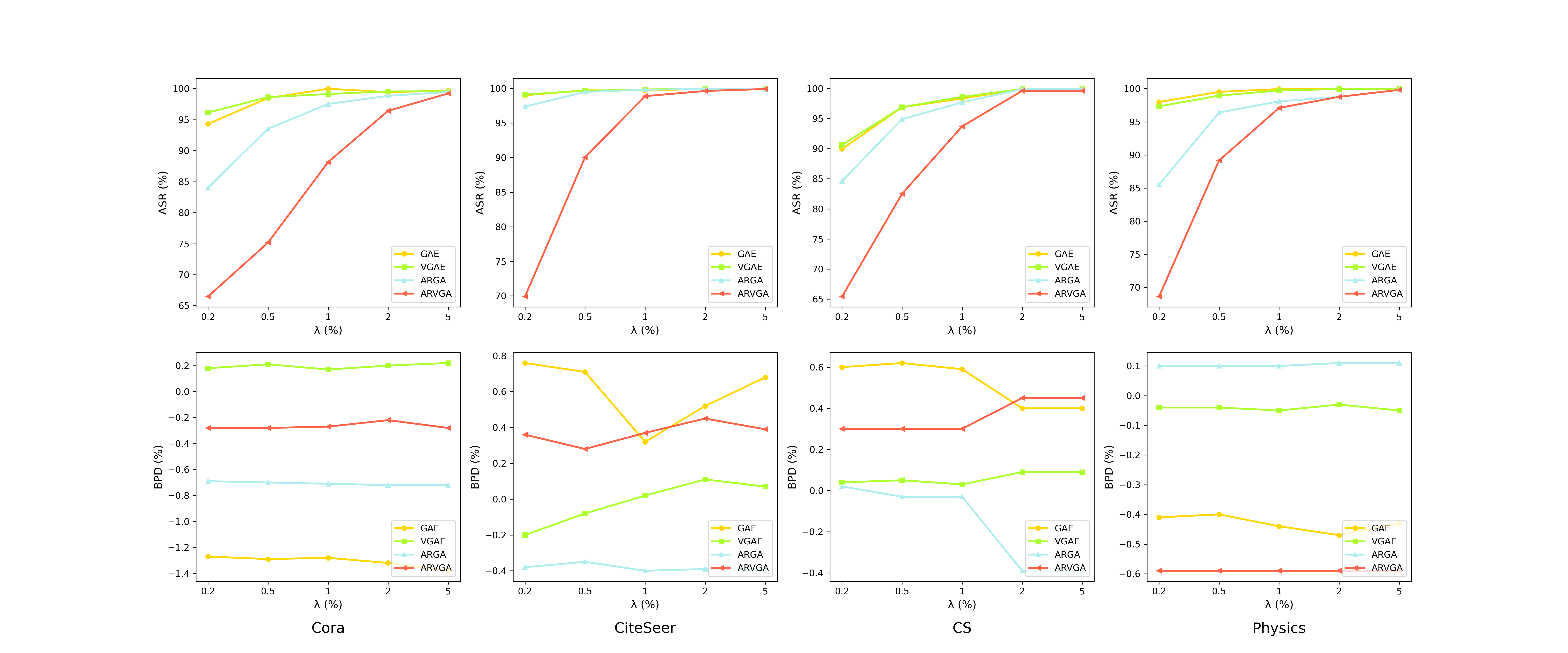}
\caption{\label{fig:size}The impact of trigger size on ASR and BPD.}
\end{figure*}

\subsection{Testing the impact of different trigger sizes on ASR and BPD}
\label{section:5.4}
\noindent In this section, we evaluate the impact of different trigger sizes on ASR and BPD of the proposed backdoor attack. As described in section \ref{section:5.1.4}, $k = \lambda d$, where $k$ is the trigger size, which represents the number of the modified elements in the feature vector of the trigger node in the process of generating the trigger node, $d$ is the total number of elements in the trigger node feature vector, and $\lambda$ is the budget, which represents the percentage of the number of modified elements in the feature vector of the trigger node compared to the total number of the elements. We evaluate the impact of diﬀerent trigger sizes on ASRs and BPDs of four backdoored models of the link predictions models (GAE, VGAE, ARGA and ARVGA) on four datasets (Cora, CiteSeer, CS, and Physics) respectively, and the values of the budget $\lambda$ in each test were 0.2\%, 0.5\%, 1\%, 2\%, and 5\%, respectively, which corresponds to the change in trigger size. The experimental results are shown in table \ref{tab:6}.

Figure \ref{fig:size} is a visualization of the experimental results, where the top four charts show the variation of ASRs with the budgets for the four backdoored models on the four datasets, with the vertical coordinate being ASR and the horizontal coordinate being $\lambda$. The bottom four charts show the variation of BPDs with the budgets for the four backdoored models on the four datasets, with the vertical coordinate being BPD and the horizontal coordinate being $\lambda$ as well. We can see from the figure that the ASRs of four backdoored models(GAE,VGAE,ARGA,ARVGA) on four datasets( Cora, CiteSeer, CS, Physics) increase as $\lambda$ increases. The backdoored models of GAE, VGAE and ARGA achieve a ASR greater than 84\% on these four datasets when $\lambda$ is just 0.2\%, while the ASR of the backdoored model of ARVGA on these four datasets is greater than 64\% and less than 70\% when $\lambda$ takes the same value,and the ASR of these four backdoored models on these four datasets achieve ASRs greater than 96\% when $\lambda$ is 2\%. The above experimental results indicate that the proposed backdoor attack can achieve good attack performance with small trigger size. Besides, we can also see from the figure that the BPDs of these four backdoored models on these four databases ﬂuctuate with the budget $\lambda$, but the fluctuations are less than 1\%, which means the trigger size has little impact on the BPDs of these backdoor models and the performance of the backdoored models on clean samples is close to that of corresponding clean models, indicating better stealthiness of the proposed backdoor attack.

\begin{figure}[!t]
\centering
\includegraphics[width=1\linewidth]{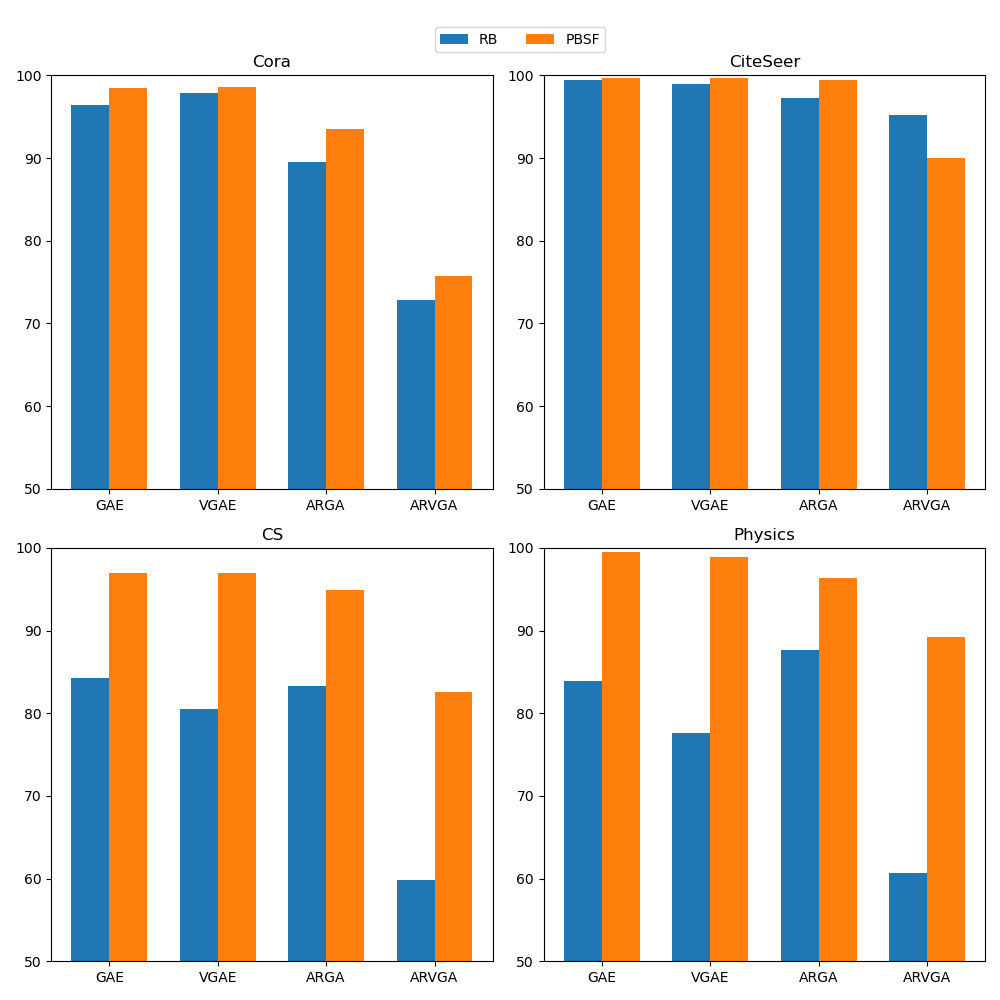}
\caption{\label{fig:Ablation Studies}The comparison of the ASR of the backoored models obtained by adopting two poisoning methods PBSF and RB.}
\end{figure}

\subsection{Ablation Studies}
\label{section:5.5}

\noindent In this section, we will conduct ablation study to confirm whether the method to generate poisoning training dataset through selecting unlinked node pairs with sparse features based on the score function to inject the trigger node(the poisoning method is called Poisoning Based on Score Function for short, PBSF), as described in section 4.3, is crucial to improving the performance of the proposed backdoor attack. For comparison, we firstly replace PBSF with another method to generate poisoning training datasets ,which randomly selects unlinked node pairs to inject the trigger(the poisoning method is called Random Poisoning for short, RB). We poison four training datasets (Cora, CiteSeer, CS and Physics) with the RB method and get the four backdoored models(GAE, VGAE, ARGA and ARVGA) with these poisoning training datasets. Then we test the ASRs of these four backdoored models on four datasets respectively and compare the results with that of the backdoored models obtained by the PBSF method. The poisoning rate is set to 1\%. The budget is 0.5\%. The experimental results are shown in figure \ref{fig:Ablation Studies}, where horizontal coordinates represents four backdoored models obtained by adopting the poisoning methods of PBSF and RB ,the vertical coordinates represent ASRs of these backdoored models, the blue bars and the orange bars in the figure correspond to the ASRs of the backdoored models generated by poisoning methods RB and PBSF respectively. From the figure, we can see that the ASRs of the backdoored models obtained by using PBSF method are higher than those of the backdoored models obtained by using RB method, especially on large dataset CS and Physics, indicating that the poisoning method PBSF is necessary for the proposed backdoor attack to achieve excellent performance.
We can also see from the figure that the ASRs of the backdoored models using the two poisoning methods are close to each other on small datasets Cora and CiteSeer. The reason for that is that most of the unlinked node pairs in the small datasets Cora and CiteSeer have sparse features,so most unlinked node pairs randomly selected by RB to inject the trigger node have sparse features with high probability, which achieve similar poisoning to PBSF that selects unlinked node pairs with sparse features by using the score function. Therefore the ASRs of the backdoored models using RB is close to that of the backdoored models using PBSF on small datasets Cora and CiteSeer. In contrast, for large large datasets CS and Physics, where most of the unlinked node pairs do not have sparse features, most unlinked node pairs randomly selected by RB to inject the trigger node are unlikely to have sparse features, and thus the ASRs of the backdoored models using RB is inferior to that of the backdoored models using PBSF.

\section{Conclusion}
\label{section:6}

\noindent This paper discusses the feasibility of implementing backdoor attacks in link prediction. Specifically, we implement a single-node injection backdoor attack that generate the features of the trigger node based on the differences in feature distribution in the dataset and identifies suitable poisoning node pairs based on the sparsity of feature pairs in the dataset. Our empirical evaluations on four representative link prediction models (GAE, VGAE, ARGA, and ARVGA) and four benchmark datasets demonstrate that our backdoor attack can achieve high ASRs with low poisoning rates, and has minimal impact on the accuracy of clean test graphs. Currently, there is limited research on defenses against backdoor attacks in graphs, and in the future, we will explore defense methods against backdoor attacks in link prediction.

\section*{Acknowledgement}
\noindent The research of the paper was supported by Natural Science Foundation of Shanghai Municipality (Grant NO.22ZR1422600)

\bibliographystyle{unsrt}
\bibliography{sample}

\end{document}